\documentclass{article}

\usepackage{arxiv}
\usepackage{fancyhdr} 
\usepackage{algpseudocode}
\usepackage[utf8]{inputenc} 
\usepackage[T1]{fontenc}    
\usepackage{hyperref}       
\usepackage{url}            
\usepackage{booktabs}       
\usepackage{amsfonts}       
\usepackage{nicefrac}       
\usepackage{microtype}      
\usepackage{lipsum}
\usepackage{graphicx}
\usepackage{hyperref}       
\usepackage{url}            
\usepackage{booktabs}       
\usepackage{amsfonts}       
\usepackage{nicefrac}       
\usepackage{microtype}      
\usepackage{xcolor}         
\usepackage{amsmath}
\usepackage{multirow}
\usepackage{algorithm}
\usepackage{amssymb}
\usepackage{mathtools}
\usepackage{wrapfig}
\usepackage{booktabs}    
\usepackage{subcaption}
\usepackage{comment}        
\graphicspath{ {./images/} }

\renewcommand{\today}{}  
\date{}                  

\usepackage{fancyhdr}
\makeatletter
\patchcmd{\@maketitle}
  {\today}                       
  {\textsc{A Preprint}}{}{}      
\makeatother

\title{Enhancing VICReg: Random-Walk Pairing for Improved Generalization and Better Global Semantics Capturing}

\author{
  Idan Simai \\
  Department of Computer Science\\
  Bar-Ilan University\\
  Ramat-Gan, Israel \\
  \texttt{idansi98@gmail.com}
  \And
  Ronen Talmon \\
  Electrical and Computer Engineering\\
  Technion\\
  Haifa, Israel\\
  \texttt{ronen@ee.technion.ac.il}
  \And
  Uri Shaham \\
  Department of Computer Science\\
  Bar-Ilan University\\
  Ramat-Gan, Israel\\
  \texttt{uri.shaham@biu.ac.il}
}

\begin{document}
\maketitle


\begin{abstract}
In this paper, we argue that viewing VICReg-a popular self-supervised learning (SSL) method--through the lens of spectral embedding reveals a potential source of sub-optimality: it may struggle to generalize robustly to unseen data due to overreliance on the training data. This observation invites a closer look at how well this method achieves its goal of producing meaningful representations of images outside of the training set as well. Here, we investigate this issue and introduce SAG-VICReg (Stable and Generalizable VICReg), a method that builds on VICReg by incorporating new training techniques. These enhancements improve the model's ability to capture global semantics within the data and strengthen the generalization capabilities. Experiments demonstrate that SAG-VICReg effectively addresses the generalization challenge while matching or surpassing diverse state-of-the-art SSL baselines. Notably, our method exhibits superior performance on metrics designed to evaluate global semantic understanding, while simultaneously maintaining competitive results on local evaluation metrics. Furthermore, we propose a new standalone evaluation metric for embeddings that complements the standard evaluation methods and accounts for the global data structure without requiring labels--a key issue when tagged data is scarce or not available.
\end{abstract}


\section{Introduction}
Self-supervised learning has emerged as a transformative paradigm in computer vision, achieving remarkable success through two primary strategies: \textbf{masking-based} methods (e.g., \cite{he2022masked,assran2022masked, assran2023learning, garrido2024learning}), and \textbf{invariance-based} methods (e.g., \cite{chen2020exploring, ermolov2021whitening, gidaris2021obow, tian2020goodviews, chen2020simple, zbontar2021barlow, caron2021emerging}).
Among these invariance-based methods, VICReg (Variance-Invariance-Covariance Regularization) \cite{bardes2021vicreg} has distinguished itself by elegantly preventing representation collapse while ensuring feature decorrelation and invariance to augmentations. While VICReg's approach has proven highly effective for representation learning, examining it through the lens of spectral embedding may reveal a possible source of sub-optimality that will be explored in this paper. 

VICReg’s approach can be understood as a form of spectral embedding on a particular graph architecture, wherein each image corresponds to a cluster and its augmentations to points within that cluster (see Section \ref{vicregasse} and Appendix \ref{sec:preliminaries}).
Like Laplacian Eigenmaps \cite{belkin2003laplacian}, VICReg uses decorrelation and invariance to align with a data‑graph’s spectral structure while avoiding collapse. While \cite{balestriero2022contrastive} initially observed this spectral connection, we extend this insight by demonstrating its potential  implications for generalization capabilities.

\begin{figure*}[!t]
\begin{tabular}{cccc}
\includegraphics[width=0.18\textwidth]{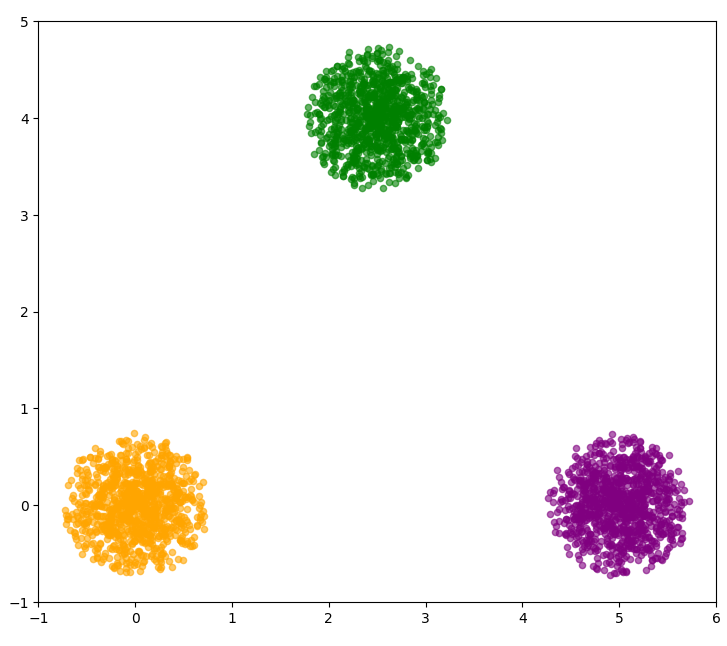} &
\includegraphics[width=0.18\textwidth]{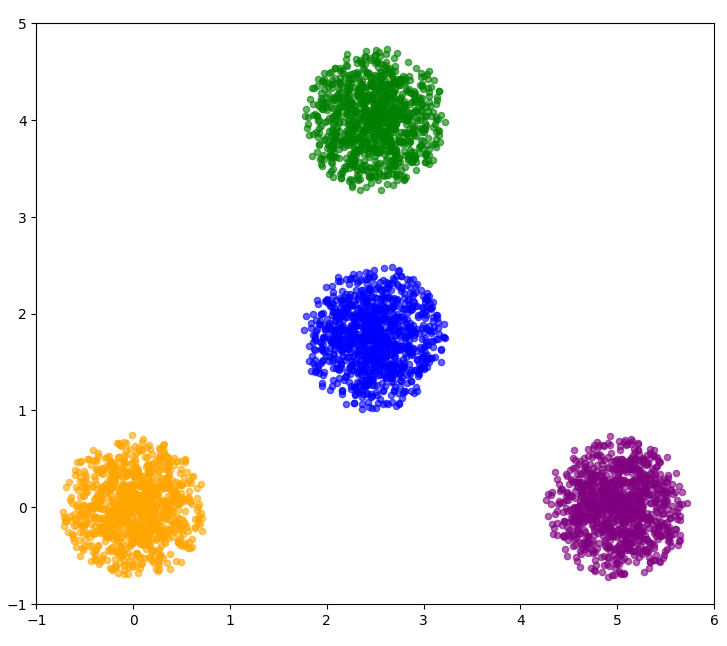} &
\includegraphics[width=0.22\textwidth]{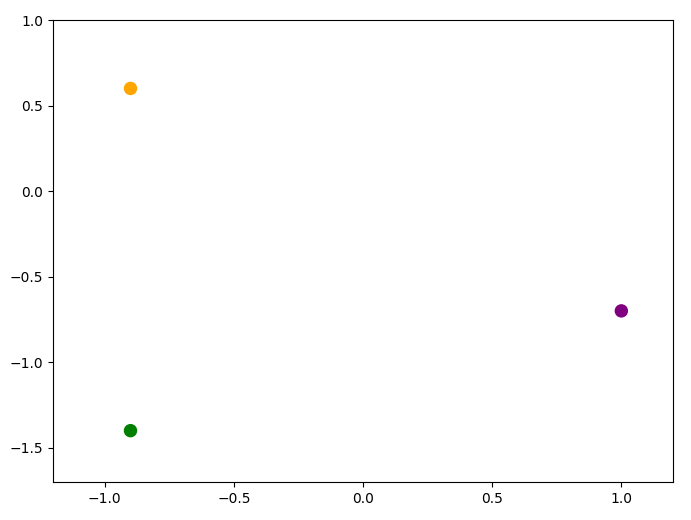} &
\includegraphics[width=0.22\textwidth]{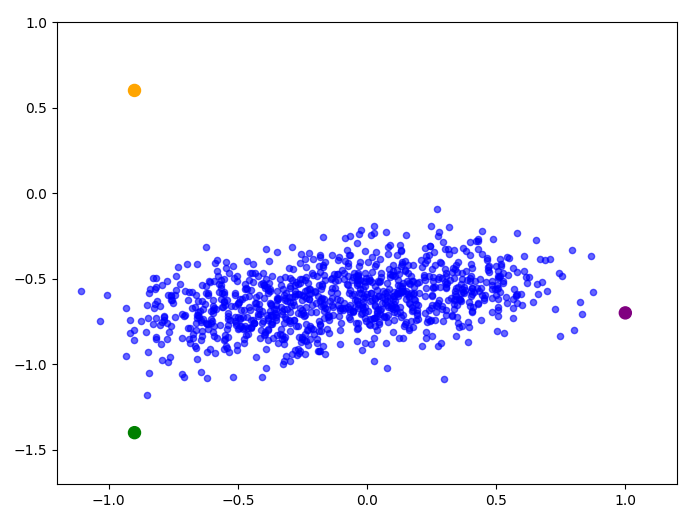} \\
(a) train points & (b) test points & (c) train points' embedding & (d) test points' embedding
\end{tabular}
\caption{\textbf{A real demonstration of the unpredictable behavior of spectral embeddings when encountering new points from an unseen cluster during 
the test}. In particular, Subfigure (a) shows the original training data with 3 clusters (green, purple, orange). Subfigure (b) displays the test data with the same 3 clusters plus a new blue cluster. Subfigure (c) shows how each of the original training clusters is embedded to a point. Subfigure (d) reveals how the embeddings of points from the new blue cluster that was added during the test show significant distortion, compared to the clear structure observed in the embeddings of the points from the original clusters.  This underscores the challenge of extending spectral embeddings to points from new, unseen clusters, and despite being a 2D example, it illustrates the generalization challenge that may occur in high-dimensional spaces as well.}
\label{fig:1}
\end{figure*}

Despite the strengths of spectral embedding methods in capturing manifold structure (e.g.,\cite{von2007tutorial, coifman2006diffusion, ng2002spectral}), typically, they are not designed to generalize. These techniques, rooted in the manifold learning domain, are specifically constructed to embed data from a predetermined graph structure. However, they are not inherently designed to generalize and produce embeddings for points that exist outside the original graph. These methods rely on the graph Laplacian’s eigenvectors (see Appendix \ref{spectralembedding}), so as long as they are not extended to eigenfunctions of the corresponding Laplacian operator, they lack an inherent mechanism to generalize to new points. VICReg, as a general embedding model, can be seen as a method aiming to generalize spectral embedding to new points.
While this approach tends to work well for data from the same clusters observed during training (as shown by SpectralNet \cite{shaham2018spectralnet}), 
it may behave unexpectedly when encountering points from new clusters during test time. This unexpected behavior is illustrated in Figure \ref{fig:1}, which shows a significant distortion in the embeddings of new points outside of the original graph's clusters, and a demonstration with real images using VICReg is provided in Figure \ref{fig:2}. This naturally raises an important question, whether VICReg optimally achieves its goal to produce meaningful embeddings of images outside of the training set as well.
This investigation forms the central focus of our manuscript--highlighting how the method can exhibit unexpected behavior when generalizing. In this paper, we provide supporting evidence for this phenomenon and propose a strategy to address the observed suboptimal performance.

We address this limitation by proposing an enhanced training strategy that extends VICReg beyond its narrow local focus, in order to improve generalization and to have more stable embeddings to new, unseen inputs. We do so by training the model to recognize not only relationships between different augmentations of the same image, but also to capture meaningful semantic relationships between distinct images across the dataset. This broader perspective enables the model to develop a richer understanding of global data structures and semantic hierarchies, allowing VICReg to effectively "escape" the boundaries of the clusters in the initial graph.

In proposing a method to improve VICReg’s stability when encountering new images and enhance its generalization, we must determine whether it truly yields higher-quality embeddings. This raises a broader question, on how image embeddings should be evaluated. Existing methods (e.g., linear or semi-supervised classification) mainly gauge the linear separability of embeddings but may be limited in their ability to capture whether the learned representations reflect the full scope of the data, as discussed in Section \ref{sec:inution}. This issue is particularly important for datasets which exhibit hierarchical structures such as CIFAR~\cite{krizhevsky2009learning} and ImageNet~\cite{deng2009imagenet}. Ideally, representations should capture both intra-class variation (e.g., pose or color changes within “vulture”) and inter-class semantics, keeping related classes closer than unrelated ones.

To overcome the limitations of existing evaluation methods, we propose a new standalone evaluation metric, which is applicable to any embedding model and accounts for the global structure of the data. This metric directly evaluates the embeddings themselves without requiring any labels–a significant advantage in scenarios where labeled data is scarce or unavailable. Additionally, it enables us to
assess the stability of any model when encountering new, unseen data. It also demonstrates that our techniques indeed effectively enhance this stability.

\section{Related Work}
\textbf{Generalization in spectral based methods and SSL.}
Traditional spectral embedding techniques such as Laplacian Eigenmaps \cite{belkin2003laplacian}, Diffusion Maps \cite{coifman2006geometric}, and their various extensions (e.g., \cite{bengio2004out, fowlkes2004spectral}) are primarily designed for fixed datasets and are not intended to generalize to new points. While approaches like the Nyström method \cite{drineas2005nystrom} or manifold-based interpolation \cite{bengio2004out} partially address this problem, they rely on carefully constructed auxiliary mappings and assume proximity between the new and the original data. More recent effort in deep spectral embedding, SpectralNet \cite{shaham2018spectralnet} focuses on scaling spectral methods to large datasets, but also aims to tackle the problem of generalization, that the traditional methods do not address. While SpectralNet excels at generalizing to new points, it assumes that the new points come from the clusters that were present during training, so embedding points of new clusters is out of its scope. Likewise, certain graph-evolution networks \cite{etemadyrad2021deep} handle dynamically changing graphs, but they are specialized to graph-structured domains and do not directly address large-scale, unlabeled image collections. In contrast, we propose an approach tailored to self-supervised visual representation learning, ensuring stability and meaningful embeddings when presented with entirely new samples.

In the broader SSL context, various methods extend beyond single-image augmentations to capture richer representations. CMC~\cite{tian2020contrastive} maximizes mutual information across different scene views, SwAV \cite{caron2020unsupervised} employs online clustering for cross-image semantic assignments, and NNCLR \cite{dwibedi2021nnclr} uses nearest neighbors as additional positive pairs. While these approaches make progress toward
capturing broader relationships, they rely on specialized architectures or complex training procedures. In contrast, our approach directly incorporates cross-image relationships through a simple yet effective framework, without using memory banks or complex training procedures.

\textbf{Evaluation of Image Embeddings.}
Traditional evaluation methods for self-supervised visual representations have primarily relied on linear or \emph{k}-NN classification protocols. Recent works have recognized the need to evaluate broader structural relationships in the learned representations. \cite{valmadre2022hierarchical}, \cite{7454509}, \cite{silla2010survey}, \cite{gordon1996hierarchical}, \cite{giunchiglia2020coherent} and \cite{cerri2014hierarchical} introduced a multi-level evaluation framework that captures both fine-grained and coarse relationships. \cite{kosmopoulos2015evaluation} provided a theoretical foundation for evaluating hierarchical relationships, unifying various approaches for measuring global data structure. While these methods successfully capture broader semantic relationships, they are inherently tied to predefined hierarchies and known class relationships. 
Our evaluation framework builds upon these insights while addressing the challenge of measuring both hierarchical structure and stability when encountering new data, without requiring explicit hierarchical annotations or ground truth labels.

\section{Preliminaries}
\label{sec:pre}
In this section we introduce SpectralNet and VICReg, and in Section \ref{sec:approach} we establish their equivalence.
\subsection{SpectralNet}
SpectralNet \cite{shaham2018spectralnet} is a deep learning approach to spectral clustering, which is intended to tackle the scalability and generalization ability issues traditional spectral embedding methods lack. The loss of SpectralNet is defined as:
\begin{equation} \label{eq:snloss}
L_{\text{SpectralNet}}(\theta) = \frac{1}{n^2} \sum_{i,j=1}^{n} W_{i,j} \| y_i - y_j \|^2_2,
\end{equation}
where \(n\) is the batch size, \( y_i \in \mathbb{R}^k \), \( y_i = F_\theta(x_i) \), \(k\) is the embedding dimension, \(y_i\) is the learned spectral embedding of the data point \(x_i\), corresponding to the coordinates of the point \(i\) in the subspace of the \(k\) leading Laplacian eigenvectors. The map \(F\) is a neural network with parameters  \(\theta\) and \( W \) is a \( n \times n \) matrix such that \( W_{i,j} = w(x_i, x_j) \), where \(w\) is a kernel function that quantifies the similarity between the points \(x_{i}, x_{j}\). The loss is computed subject to an orthogonality constraint which is enforced using an orthogonalization layer in the architecture:
\begin{equation} \label{eq:orthogonality}
    \frac{1}{n} Y^T Y = I_{k \times k,}
\end{equation}
where \( Y \) is a \( n \times k \) matrix of the outputs whose \( i \)-th row is \( y_i^T \).
SpectralNet is a spectral embedding method because the matrix \(Y\), which minimizes the SpectralNet loss \eqref{eq:snloss} while satisfying the orthogonality constraint \eqref{eq:orthogonality}, corresponds to the leading eigenvectors of the Laplacian \(L = D -W\), where \(D\) is the degree matrix, as explained in Appendix \ref{spectralembedding}. Minimizing \eqref{eq:snloss} is equivalent to minimizing: $
\text{trace} \left( Y^T (D - W) Y \right) = \text{trace} \left( Y^T L Y \right)$.
Hence, SpectralNet functions as a spectral embedding technique.
As it is a parametrized map (by a neural network), SpectralNet has the ability to generalize to new points during inference. 

\subsection{VICReg}
VICReg is an SSL method designed to produce robust embeddings by imposing specific regularization terms. The invariance term makes sure that the model is invariant to different augmentations, while the covariance and variance terms prevent a collapse (where the model outputs constant vectors).\\
\textbf{Variance Regularization:} This term ensures adequate spread in embeddings along each dimension. For a batch of embeddings \( Y = [y_1, y_2, \dots, y_n] \), define \(y^j=(y_1^j,\ldots,y_n^j)^T\). The term is:
\(
    v(Y) \;=\; \frac{1}{k} \sum_{j=1}^k \max\!\bigl(0,\, \gamma \;-\; S\!\bigl(y^j, \epsilon\bigr)\bigr),
\)
   where \( S \) is the regularized standard deviation defined by:
\(
S(x, \epsilon) = \sqrt{\text{Var}(x) + \epsilon},
\)
   \( k \) is the embedding dimension, \( \gamma \) is a target threshold and \( \epsilon \) is a small constant.\\
\textbf{Invariance Regularization:} This term promotes consistency between different augmentations of the same image, effectively encourages semantically closer images to have similar embeddings. For two sets of embeddings \( Y \) and \( Y' \) of augmented views, the term is:
\(
   s(Y, Y') = \frac{1}{n} \sum_{i=1}^n \| y_i - y_i' \|_2^2
\).\\
\textbf{Covariance Regularization:} This term minimizes redundancy by reducing correlations among embedding dimensions. For a batch of embeddings \( Y \), the overall term is:
\(
c(Y) = \frac{1}{k} \sum_{i \neq j} \left[C(Y)\right]_{i,j}^2,
\)
where \(C(Y)\) is the sample covariance matrix of \(Y\). The overall loss function in VICReg is a weighted combination of the invariance, variance, and covariance terms.

\section{Proposed approach}
\label{sec:approach}
\subsection{VICReg as a Spectral Embedding Method}
\label{vicregasse}
Before introducing our techniques and the rationale behind them, we first establish the fundamental connection between VICReg and spectral embedding, following the mathematical introduction provided in Appendix \ref{sec:preliminaries} and Section \ref{sec:pre}. This connection becomes apparent when we examine the losses of SpectralNet and VICReg. In VICReg, the invariance term can be defined as:
\begin{equation}
\label{weightedinv}
    s(Y, Y') = \frac{1}{n} \sum_{i,j=1}^n W_{ij} \| y_i - y_j' \|_2^2,
\end{equation}
where \( W_{ij} = 1 \) if \( i = j \), meaning \( y_i \) and \( y_j' \) are embeddings of augmentations of the same image, and \( W_{ij} = 0 \) otherwise. This is precisely the same expression used in SpectralNet's loss \eqref{eq:snloss}, but in the specific case where \( W_{ij} \in \{0, 1\} \), whereas in SpectralNet, \( W_{ij} \in (0,1] \). Additionally, when we consider the covariance term combined with the variance term in VICReg, we effectively obtain the identity covariance constraint\footnote{Under the assumption of zero mean of the vectors \(y_{i} \in Y\), but the Laplacian's eigenvectors are orthogonal to the constant vector (the first eigenvector), ensuring that the mean from the second and on is zero.}
which is exactly the same orthonormalized output constraint \eqref{eq:orthogonality} as in SpectralNet. These observations allow us to view VICReg as a specific instance of SpectralNet, considering a special case of a graph. This graph is structured in clusters in which each cluster is created by an image and each node within that cluster is one of its augmentations. An edge (with weight 1) connects two nodes only if they are augmentations of the same image; otherwise, the edge weight is 0.
Therefore, VICReg is also a spectral embedding method, just like SpectralNet.

Recognizing this connection to spectral embedding, we identify a potential unexpected behavior.
VICReg is a spectral embedding method that aims to produce meaningful embeddings even for images from outside the training clusters, a scenario for which such approaches (including the mathematically equivalent SpectralNet) are not inherently designed to deal with. This raises the question whether VICReg optimally behaves as it is excepted to.

\begin{figure*}[h!]
    \centering
\includegraphics[width=1\textwidth]{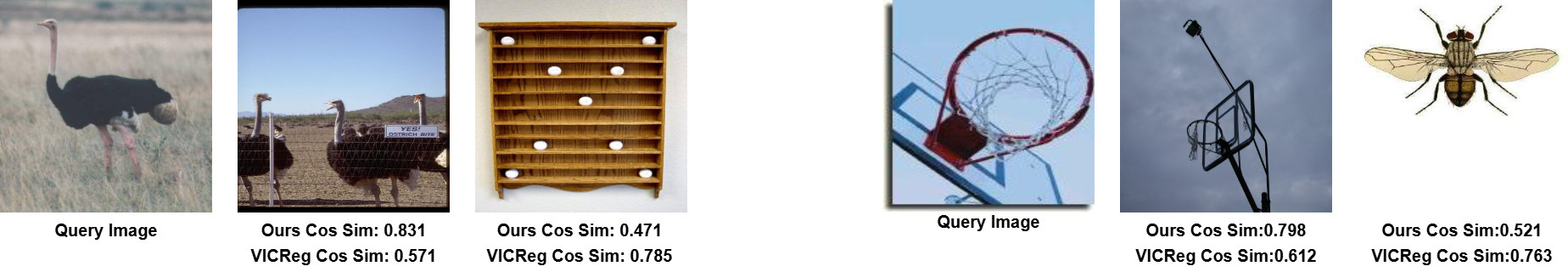} 
    \caption{\textbf{A real-world demonstration of how VICReg behaves unstably when encountering images from unseen clusters during test time.} We compare cosine similarity scores between the query image (leftmost) and other images, showing results from both VICReg and our approach. Notice how VICReg’s embeddings can place semantically similar images far apart and semantically unrelated images too close together—an empirical demonstration of the unexpected behavior.}
    \label{fig:2}
\end{figure*}
Figure \ref{fig:2} provides a real-world demonstration of the embedding inconsistency we identified, directly paralleling the 2D illustration in Figure \ref{fig:1}. This empirical demonstration aligns with the findings in Figure \ref{fig:1}, in that ostrich images are embedded in a manner lacking meaningful structure--so that a wooden shelf appears closer in the embedding space than other ostrich (just like the points from the blue cluster in (d) that were not present during training, whose points are much dispersed). Alongside these illustrations, Figure \ref{fig:8} in Appendix \ref{sec:visual} provides UMAP visualizations comparing our approach to VICReg when embedding images from unseen classes. These visualizations highlight our approach's ability to maintain coherent semantic structure when generalizing to new, unseen data, while VICReg's embeddings exhibit noticeable distortions and inconsistencies.
\subsection{Rationale and Intuition}
\label{sec:inution}
We now describe the rationale behind SAG-VICReg and the modifications we propose to improve VICReg's generalization and stability to new, unseen data. Alongside these modifications, we will give the motivation and illustrate why metrics like linear classification may be limited in capturing whether the learned representations reflect the full scope of the data, and how our method overcomes it. In addition, we provide a theoretical justification to our approach in Appendix \ref{sec:theory_related}.

To address the problem of generalization and the instability of VICReg, we design an algorithm that conducts modifications to the training process that improves out-of-cluster generalization, and aims to push VICReg beyond its initial cluster dependencies. Our approach achieves this by adjusting the batch structure to not include only pairs of augmentations of the same image (as in the original VICReg setup), but also pairs of augmentations of different images that are "close" in some sense. To further enhance this effect, we weight the loss function according to the similarity of each pair. This out-of-cluster approach encourages the model to produce embeddings that are more stable and meaningful even for new, unseen data, and to learn broader relationships between data points.

As discussed earlier, we propose a new metric that goes beyond linear separability to evaluate the structural integrity of embeddings, \emph{without requiring any labels}. The rationale is that linear-separability scores tell only part of the story--an embedding can preserve the data’s global strcture even when classes are not cleanly split by a linear probe, as the top row of Figure \ref{fig:3} in Appendix \ref{sec:visual} shows.
To capture this broader quality notion, we introduce a \emph{label-free} metric that evaluates embeddings directly.
Rather than focusing solely on immediate neighbors like \emph{k}-NN, it analyses relationships across the entire dataset, revealing coarse-to-fine semantic structure. As it is computed on the representations themselves without any downstream task or ground-truth labels, it offers a direct, \emph{label-free} measure of structural integrity, as can be seen in The bottom row of Figure \ref{fig:3}.

\subsection{SAG-VICReg}
Our approach initially follows the VICReg procedure up to the point where two batches of embeddings are produced.
Let \( S = \{s_1, s_2, \ldots, s_n\} \subset \mathcal{D} \) be a mini-batch of images sampled from dataset \( \mathcal{D} \). For each image \( s_i \in S \), we apply two random transformations \( t, t' \sim \mathcal{T} \), where \( \mathcal{T} \) is a set of augmentations.
The transformed views \( x \) and \( x' \) are then passed through a SAG-VICReg network (a backbone model as in VICReg). First, they are encoded by the feature extractor \( f_\theta \) (which is the backbone, in VICReg and in our SAG-VICReg, it is a ResNet \cite{He2016} or a ViT \cite{dosovitskiy2020image} backbone) into representations \( y = f_\theta(x) \) and \( y' = f_\theta(x') \). These representations are then mapped by an expander \( h_\phi \)  (which is a Neural network consisting of fully-connected layers) to the embeddings \( z = h_\phi(y) \) and \( z' = h_\phi(y') \), which will be used to compute the loss. The embeddings \( Z = \{z_1, z_2, \ldots, z_n\} \) and \( Z' = \{z'_1, z'_2, \ldots, z'_n\} \) form two batches, for further processing.

\begin{wrapfigure}{r}{0.3\textwidth}  
\vspace{-1em}
\centering
\includegraphics[width=\linewidth]{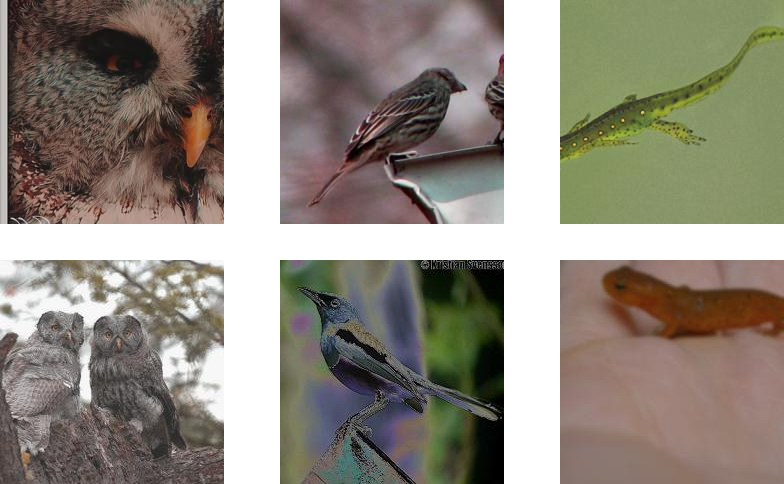}
\vspace{-12pt}
\caption{\textbf{Non-trivial pairs.} For each column, the upper image shows where the random walk starts and the lower image shows where it ends.}
\label{fig:4}
\end{wrapfigure}

At this stage, we introduce our primary contribution by incorporating relationships among the embeddings.
In order to capture relationships between the embeddings, we construct an affinity matrix \( W \in \mathbb{R}^{n \times n} \) based on cosine similarity. For each sample \( z_i \in Z \), we identify its \( k \)-nearest neighbors in \( Z' \) and compute the cosine distance \( d_{ij} = 1 - \cos(z_i, z'_j) \) between \( z_i \) and each of its neighbors \( z'_j \). We define each entry of \( W \) using a Gaussian kernel:
$
W_{ij} = \exp\left(-\frac{d_{ij}^2}{\sigma_i^2}\right),
$
where \( \sigma_i \) is a local scaling factor meaning its value varies for each individual point.
Implementation details including local scaling factor calculations are provided in Appendix \ref{implementation}, with hyperparameters considerations in Appendix \ref{hyperparameters}. SAG-VICReg remains stable across different hyperparameter configurations.
Next, we construct a Random Walk matrix \( P\) and sample pairs as in Appendix \ref{randomwalk}.
Some examples of non-trivial pairs generated through this sampling are shown in Figure \ref{fig:4}. 
This sampling process creates a new batch of pairs \( (z_i, z''_i) \), where \( z''_i \) is selected from \( Z' \) using the the Random Walk \(P\) starting from \( z_i\).
Finally, we compute the invariance term in the VICReg loss, weighted by the similarity values \( W_{ij} \) \eqref{weightedinv}. Specifically, the invariance constraint is adjusted by weighting each pair \( (z_i, z''_i) \) according to \( W_{ij} \). The remaining terms in the VICReg loss function remain unchanged. Our random‑walk pairing can be viewed as a form of \emph{nearest‑neighbour debiasing}, similar to the neighbour‑propagation step used in I-Con \cite{alshammari2025icon}. 
The key difference is motivational: I‑Con debiases to correct false‑negative noise from random negative sampling, whereas our goal is to bolster out‑of‑cluster generalization.
The full pseudo code is described in Algorithm \ref{alg:alg1}.

\vspace{0.2em} 
\begin{algorithm}[H]
\caption{SAG-VICReg}
\label{alg:alg1}
\textbf{Input:} Mini-batch of images \( S = \{s_1, s_2, \ldots, s_n\} \subset \mathcal{D} \), transformations \( t, t' \sim \mathcal{T} \)\\
\textbf{Output:} None
\begin{algorithmic}[1]
\State Apply transformations \( t \) and \( t' \) to \( S \), generating two transformed views \( x = t(i) \) and \( x' = t'(i) \)
\State Forward the views \( x \) and \( x' \) through the network by \(z = h_\phi(f_\theta(x)) \), \( z' = h_\phi(f_\theta(x')) \)
and form 2 batches
\( Z = \{z_1, \ldots, z_n\} \) and \( Z' = \{z'_1, \ldots, z'_n\} \) 
\State Compute affinity matrix \( W \in \mathbb{R}^{n \times n} \) between samples in \( Z \) and \( Z' \)
\State Compute a Random Walk matrix \( P\) and sample \(n\)
pairs to form the new batch \( Z'' = \{z_1'', \ldots, z_n''\} \), where every \( z''_i \) is selected from \( Z' \)
\State Compute the weighted VICReg's invariance loss using \( W_{ij} \) between \(Z\), \(Z''\) \eqref{weightedinv} and compute the regular VICReg's variance and covariance on each \(Z\), \(Z''\) separately
\end{algorithmic}
\end{algorithm}
\vspace{-2.0em} 
\subsection{Global Structure‐Preserving Embedding Evaluation}
\label{sec:hier}
We introduce a novel \emph{label-free} evaluation metric designed to capture both the hierarchical and global structure preservation of embeddings, as well as the generalization capabilities of embedding models. This is a stand-alone contribution, useful for assessing embedding models in general.
Let \( A = \{ a_1, a_2, \dots, a_n \} \subset \mathbb{R}^d \) and \( B = \{ b_1, b_2, \dots, b_n \} \subset \mathbb{R}^m \)
be two sets of embeddings, each containing \(n\) vectors, where \(a_i \in \mathbb{R}^d\) and \(b_i \in \mathbb{R}^m\) are embeddings of the same entity for \(i = 1, \dots, n\). The dimensions \(d\) and \(m\) may or may not be equal. These sets might come from two different models or represent original data alongside its embeddings. We use these two sets to construct hierarchical trees (dendrograms), and evaluate their structural similarity, capturing how well the global organization of one embedding space corresponds to the other. Using this metric, we can assess the generalization capabilities of models. For example, if \( A\) represents embeddings generated by a model on data it was trained on and \( B \) represents embeddings of data the model has not seen before, high similarity between their dendrograms indicates that the model has effectively captured the underlying data relationships, even in unseen data.

\paragraph{Hierarchical Tree Construction.}
To construct the hierarchical trees for \(A\) and \(B\), we use an agglomerative clustering algorithm \cite{mcquitty1957elementary} with cosine distance. We employ Ward linkage \cite{anderberg1973cluster}, which minimizes within-cluster variance, creating compact clusters and a clear hierarchy.

\paragraph{Lowest Common Ancestor (LCA) Similarity.}
Once the dendrograms are built, we quantify similarity by computing the distance of every embedding pair to their \emph{lowest common ancestor} (LCA) in each tree. A smaller LCA distance indicates a closer relationship within the dendrogram. We collect these LCA distances for all pairs from both dendrograms, then calculate their Pearson, Spearman, and Kendall correlation coefficients. These correlations reflect how the global structure in the dendrograms is correlated. We refer to these correlations as the \textbf{LCA similarity}.

\paragraph{Cophenetic Similarity.}
We also compute the \emph{cophenetic correlation} coefficient (see Appendix \ref{cope}), which evaluates how well each set’s \emph{cophenetic distances} (dendrogram-based distances at which pairs of items join the same cluster) correspond to the actual pairwise distances in the other set. This measures how the global structure correlates the local structure between the two sets, capturing global and local similarities as well. We refer these similarities as the \textbf{Cophenetic similarity}.

\begin{algorithm}
\caption{Evaluation of Structural Similarity between Embedding Sets}
\label{alg:algo2}
\textbf{Input:} Two embedding sets \( A = \{ a_1, a_2, \dots, a_n \} \subset \mathbb{R}^d \) and \( B = \{ b_1, b_2, \dots, b_n \} \subset \mathbb{R}^m \)\\
\textbf{Output:} Similarity scores based on hierarchical and global structure preservation
\begin{algorithmic}[1]
\State Construct hierarchical trees \( T_1 \) and \( T_2 \) for \( Z_1 \) and \( Z_2 \) using agglomerative clustering, and calculate the
distances to LCA and the cophenetic distances \(D_1\), \(D_2\) for all pairs
\State Compute Pearson, Spearman, and Kendall correlation coefficients between the LCA distances of \( A \) and \( B \)  
\State Compute cophenetic correlation coefficient
    between \( D_1 \) and the pairwise distances in \( B \), \( D_2 \) and the pairwise distances in \( A \)
\State Return all the calculated correlation coefficients
\end{algorithmic}
\end{algorithm}
\vspace{-1.2em}

\paragraph{Interpretation and Generalization.}
By correlating dendrogram-based distances, our metric quantifies how well an embedding preserves the \emph{full, multi-scale hierarchy} of the data.  In Figure~\ref{fig:3} in Appendix \ref{sec:visual}, embedding~2 yields dendrograms that almost perfectly mirror those of the original points, resulting in much higher LCA and cophenetic correlations than embedding~1.  
As the computation is entirely \emph{label-free} and involves no downstream task, the metric remains applicable even when annotations are scarce or absent.  It complements local tests such as \emph{k}-NN or linear probes by capturing broader semantic relationships and hierarchical structures that exist across the entire dataset, rather than just pairwise similarity or class separability.
Additionally, this method naturally extends to measuring generalization, as it can directly assess whether a model organizes unseen data with the same structural patterns it learned during training.

\section{Experiments and Results}
\label{sec:experandresults}
We evaluate SAG-VICReg through comprehensive experiments, comparing it against VICReg, other invariance-based spectral embedding methods (Barlow Twins, SimCLR), and other state-of-the-art approaches including DINO and masking-based techniques.
\paragraph{\textbf{Datasets and Architecture.}} We use 3 well-known datasets, ImageNet, CIFAR-100 and Caltech-256 \cite{caltech256}.
These datasets are selected not only because they are well-studied and widely recognized, but also because they have clear hierarchy \cite{yan2015hdcnn, DBLP:journals/corr/RedmonF16, kim_class_hierarchies}. This makes them suitable for evaluating whether our SAG-VICReg method enhances VICReg's ability to capture global structure more effectively.
The architecture is detailed in Appendix \ref{Arch}.
\paragraph{\textbf{Setup.}}
A model with strong generalization capabilities should embed an image similarly whether or not it encountered that image (or its class) during training. To test this, we examine how the embeddings for the same images differ depending on whether the model has seen those images during training or not.
     The complete information on this experiment is provided in Appendices \ref{data} and \ref{setup}.
To do so, for a given dataset, we compare the embeddings sets generated by the two models by calculating the correlations between them using LCA and Cophenetic similarity. Higher correlation indicates better generalization ability, as explained in Section \ref{sec:hier}.

As an additional evaluation metric, we calculate the hierarchical Rand index (see  Appendices \ref{rand_index} and \ref{rand_index_setup}) at different levels of the CIFAR-100 class hierarchy. Specifically, we compare the ground-truth labels in the hierarchy with the cluster assignments of the embeddings. This metric provides insight into how effectively the model separates semantically distinct groups of classes, and clusters together semantically closer groups of classes. A higher Rand index indicates that the embedding-based clustering is more consistent with the known hierarchical labels, reflecting stronger preservation of the global data structure. We do so for different hierarchy levels (where level 1 corresponds to high-level category labels and level 4 to the finest-grained class
label). We also compute the Rand index for the finest-grained class labels with various numbers of clusters (see Appendix \ref{rand_index_setup}) to assess whether the results are sensitive to this number. 

For completeness, and despite we discussed the shortcomings of these methods, we also evaluate our model using hierarchical linear and \emph{k}-NN classification at different hierarchy levels for a more comprehensive performance assessment, using ground-truth labels.

\begin{table*}[h!]
    \centering
    \caption{LCA similarity results. Relative gains over the second-best are shown; best scores are bolded.}
    \label{tab:correlations_between_trees}
    \scalebox{0.60}{ 
    \begin{tabular}{lccc|ccc|ccc}
        \toprule
        \multirow{2}{*}{\textbf{}} & \multicolumn{3}{c}{\textbf{ImageNet 1k}} & \multicolumn{3}{c}{\textbf{CIFAR-100}} & \multicolumn{3}{c}{\textbf{Caltech-256}} \\
        \cmidrule(lr){2-4} \cmidrule(lr){5-7} \cmidrule(lr){8-10}
         & \textbf{Pearson} & \textbf{Spearman} & \textbf{Kendall} & \textbf{Pearson} & \textbf{Spearman} & \textbf{Kendall} & \textbf{Pearson} & \textbf{Spearman} & \textbf{Kendall} \\
        \midrule
        
        \textbf{SimCLR} & $0.248\pm0.04$ &
        $0.348\pm0.1$ &
        $0.296\pm0.06$ & 
        \underline{$0.235\pm0.02$} & \underline{$0.226\pm0.01$} & $0.187\pm0.05$ &
        $0.289\pm0.07$ &
        $0.262\pm0.03$ &
        $0.211\pm0.04$\\
        \textbf{Barlow Twins} & 
        $0.247\pm0.02$ &
        $0.321\pm0.02$ &
        $0.257\pm0.02$ & 
        $0.204\pm0.01$ & $0.219\pm0.06$ & $0.181\pm0.05$ &
        \underline{$0.306\pm0.02$} &
        \underline{$0.317\pm0.11$ }&
        \underline{$0.254\pm0.09$ }\\
        \textbf{VICReg} & \underline{$0.287\pm0.03$} & \underline{$0.403\pm0.06$} & \underline{$0.329\pm0.05$} &
        $0.202\pm0.01$ & $0.223\pm0.02$ & \underline{$0.193\pm0.02$} & 
        $0.266\pm0.01$ & $0.239\pm0.04$ & $0.201\pm0.04$ \\

        \textbf{SAG-VICReg (Ours)} & \textbf{\boldmath$0.318\pm0.04$} & \textbf{\boldmath$0.433\pm0.07$} & \textbf{\boldmath$0.351\pm0.05$} & 
        \textbf{\boldmath$0.268\pm0.02$} & \textbf{\boldmath$0.296\pm0.12$} & \textbf{\boldmath$0.25\pm0.09$} &
        \textbf{\boldmath$0.321\pm0.04$} & \textbf{\boldmath$0.413\pm0.08$} & \textbf{\boldmath$0.35\pm0.07$} \\
        
        \textbf{Gain} & $10.8\%$ & $6.7\%$ & $6.3\%$ & 
        $14.04\%$ & $30.03\%$ & $29.53\%$ & $4.57\%$ & $30.28\%$ & $37.79\%$ \\
        \bottomrule
    \end{tabular}
    }
    \label{table:1}
\end{table*}
\begin{table*}[h!]
    \centering
    \caption{Cophenetic similarity results.
    \textbf{D1 to P2} correlates second model's pairwise distances with first model's cophenetic distances, and vice versa for \textbf{D2 to P1}. Relative gains over the second-best are shown; best scores are bolded.}
    \scalebox{0.60}{
    \begin{tabular}{lcc|cc|cc}
        \toprule
        \multirow{2}{*}{\textbf{}} & \multicolumn{2}{c}{\textbf{ImageNet 1k}} & \multicolumn{2}{c}{\textbf{CIFAR-100}} & \multicolumn{2}{c}{\textbf{Caltech-256}} \\
        \cmidrule(lr){2-3}\cmidrule(lr){4-5}\cmidrule(lr){6-7}
         & \textbf{D1 to P2} & \textbf{D2 to P1} & \textbf{D1 to P2} & \textbf{D2 to P1} & \textbf{D1 to P2} & \textbf{D2 to P1} \\
        \midrule
        \textbf{SimCLR} &
        \underline{$0.289\pm0.03$} & \textbf{\boldmath$0.259\pm0.06$} &
        \underline{$0.349\pm0.08$} & \underline{$0.341\pm0.09$} &
        $0.305\pm0.06$ & $0.322\pm0.09$\\
        
        \textbf{Barlow Twins} &
        $0.199\pm0.01$ & $0.156\pm0.01$ &
        $0.309\pm0.004$ & $0.286\pm0.01$ &
        \underline{$0.341\pm0.01$} & \underline{$0.364\pm0.03$}\\
        
        \textbf{VICReg} &
        $0.263\pm0.03$ & $0.213\pm0.01$ &
        $0.287\pm0.006$ & $0.277\pm0.01$ &
        $0.342\pm0.02$ & $0.335\pm0.01$\\
        
        \textbf{SAG-VICReg (Ours)} &
        \textbf{\boldmath$0.308\pm0.03$} & \underline{$0.246\pm0.02$} &
        \textbf{\boldmath$0.405\pm0.02$} & \textbf{\boldmath$0.384\pm0.03$} &
        \textbf{\boldmath$0.382\pm0.01$} & \textbf{\boldmath$0.375\pm0.03$}\\
        
        \textbf{Gain} &
        $6.57\%$ & $-5.28\%$ &
        $15.71\%$ & $12.61\%$ &
        $12.02\%$ & $3.02\%$\\
        \bottomrule
    \end{tabular}
    }
    \label{table:2}
\end{table*}
\paragraph{\textbf{Results.}}
In Table \ref{table:1}, we report the LCA similarity results across multiple datasets. SAG-VICReg consistently achieves the highest scores, with particularly notable gains in Spearman and Kendall correlations. These strong results demonstrate our method's enhanced ability to generalize to previously unseen data while preserving global semantic structure-a crucial capability that goes beyond the local neighborhood focus of traditional SSL methods. While Pearson captures linear relationships, Spearman and Kendall more accurately reflect rank correlations, which are essential for evaluating hierarchical structures since they measure how well the ordering relationships between data points are preserved rather than just their absolute distances.
Table \ref{table:2} shows the Cophenetic similarity results. Here, SAG-VICReg again demonstrates superior performance. This indicates that our approach effectively maintains both local pairwise distances and global hierarchical relationships during generalization, a challenging balance to achieve in representation learning.\\
\\
Table \ref{table:3} compares our approach with state-of-the-art masking-based models and DINO on CIFAR-100. \begin{wraptable}{r}{0.50\textwidth}  
\vspace{-1em}  
\centering
\caption{LCA similarity results. Best scores are bolded.}
\scalebox{0.62}{%
\begin{tabular}{lccc}
   \toprule
   \multicolumn{1}{c}{} & \multicolumn{3}{c}{\textbf{CIFAR-100}} \\ 
   \cmidrule(lr){2-4}
    & \textbf{Pearson} & \textbf{Spearman} & \textbf{Kendall} \\
   \midrule
   \textbf{VICReg} & 
   $0.145\pm0.05$ &
   $0.185\pm0.09$ &
   $0.152\pm0.09$\\ 
   \textbf{DINO} & 
   \underline{$0.242\pm0.04$} &
   \underline{$0.262\pm0.09$} &
   \underline{$0.213\pm0.08$}\\
   \textbf{MAE} & 
   $0.238\pm0.015$ & 
   $0.209\pm0.03$ &
   $0.174\pm0.03$ \\
   \textbf{I-JEPA} & 
   \textbf{\boldmath$0.325\pm0.04$} & 
   $0.189\pm0.08$ & 
   $0.161\pm0.07$\\
   \textbf{SAG-VICReg (Ours)} & 
   $0.201\pm0.02$ & 
   \textbf{\boldmath$0.269\pm0.05$} & 
   \textbf{\boldmath$0.221\pm0.05$}\\
   \bottomrule
\end{tabular}%
}
\label{table:3}
\vspace{-1em} 
\end{wraptable}The results reveal a remarkable transformation--our enhancements elevate VICReg from the worst-performing method (in terms of Spearman and Kendall correlations) to the best-performing approach in capturing global semantic structure while generalizing, outperforming even sophisticated architectures like DINO and I-JEPA in rank correlations. This dramatic improvement in global structure preservation while generalizing demonstrates how our random-walk pairing strategy effectively enhances the capturing of global semantics and generalization capabilities. Specifically, it improves the model's ability to generalize to unseen data while maintaining meaningful semantic relationships, without requiring architectural changes or extensive additional training complexity.\\
\begin{figure}[!h] 
\captionsetup{belowskip=-7pt} 
  \begin{subfigure}[t]{0.52\linewidth}
    \centering
    \includegraphics[width=\linewidth]{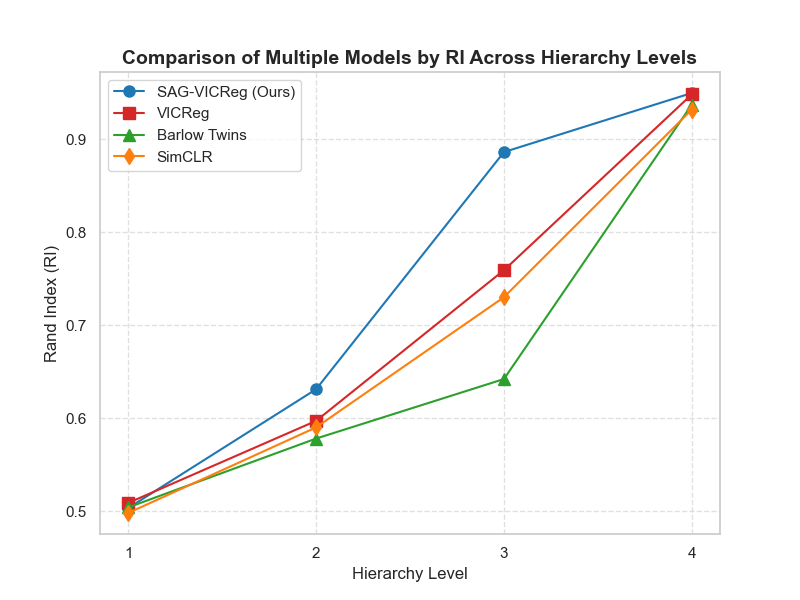}
    \caption{Rand Index vs.\ hierarchy level.}
    \label{fig:6}
  \end{subfigure}
  \hfill
  \begin{subfigure}[t]{0.52\linewidth}
    \includegraphics[width=\linewidth]{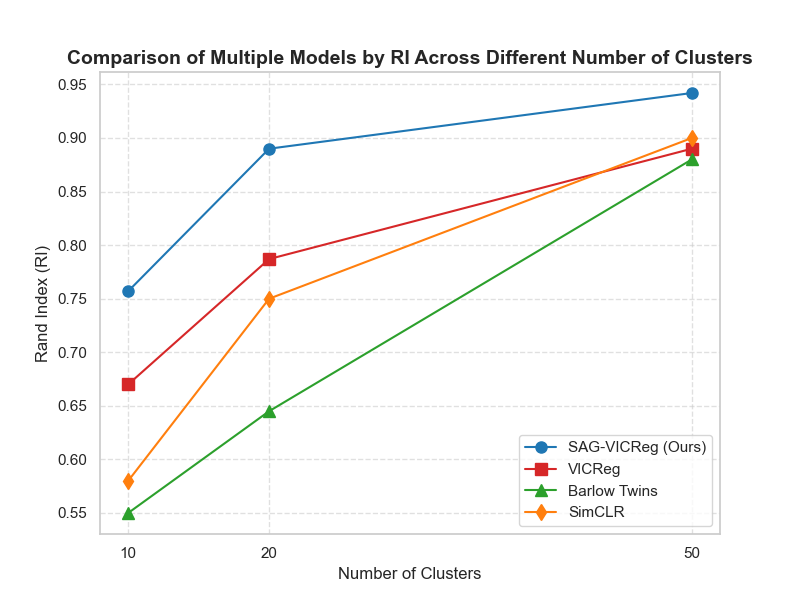}
    \caption{Rand Index vs.\ number of clusters.}
    \label{fig:7}
  \end{subfigure}
  \caption{Comparison of Rand‐Index scores across hierarchy levels (left) and varying numbers of clusters (right).}
  \label{fig:ri_comparison}
\end{figure}
\vspace{2em}
\\Figure \ref{fig:6} presents the Rand Index across different hierarchical levels of CIFAR-100. SAG-VICReg consistently achieves the highest scores across all hierarchy levels, with especially pronounced advantages at higher, more abstract levels (Levels 2 and 3). This indicates that our method not only distinguishes between fine-grained categories but also correctly organizes these categories into their broader semantic relationships, reflecting the natural hierarchical organization present in the data.
Figure \ref{fig:7} demonstrates SAG-VICReg's robustness to varying cluster counts when evaluating against the finest-grained labels. This stability is crucial for real-world applications where the optimal number of semantic groups may be unknown or may change over time. While other methods show significant performance fluctuations as the cluster count changes, SAG-VICReg maintains consistently superior performance, indicating that it captures stable and meaningful semantic structure regardless of clustering granularity.
\vspace{2em}
\begin{table*}[h!]
\caption{Hierarchical \emph{k}-NN classification on CIFAR-100}
\label{table:4}

\begin{subtable}[t]{0.35\linewidth}
\centering
\caption{\textbf{Pretraining on CIFAR-100}}
\label{tab:knn_cifar100_pretrain_cifar}
\small
\scalebox{0.62}{
\begin{tabular}{lcccc}
\toprule
\textbf{Model} & \textbf{Level 1} & \textbf{Level 2} & \textbf{Level 3} & \textbf{Level 4} \\
\midrule
\textbf{SimCLR}                     & 71.26\% ± 0.66  &  54.86\% ± 0.72 & 39.15\% ± 0.54 & 27.06\% ± 0.31\\
\textbf{BarlowTwins}   & \underline{73.17\% ± 0.45}
& 56.93\% ± 0.48 &  \underline{41.65\% ± 0.32} &  29.67\% ± 0.26 \\
\textbf{VICReg}                     & 73.01\% ± 0.51 &\underline{57.07\% ± 0.52}  & 
41.62\% ± 0.43 & 
\underline{29.86\% ± 0.19}\\
\textbf{SAG-VICReg (Ours)}          & \textbf{74.33\% ± 0.43}  & \textbf{58.25\% ± 0.39} & \textbf{41.99\% ± 0.33} &
\textbf{29.87\% ± 0.20}\\
\bottomrule
\end{tabular}}
\end{subtable}
\hspace{8.5em} 
\begin{subtable}[t]{0.35\linewidth}
\centering
\caption{\textbf{Pretraining on ImageNet-1k}}
\label{tab:knn_cifar100_pretrain_imagenet}
\small
\scalebox{0.62}{
\begin{tabular}{cccc}
\toprule
\textbf{Level 1} & \textbf{Level 2} & \textbf{Level 3} & \textbf{Level 4} \\
\midrule
83.12\% ± 0.49 & 71.75\% ± 0.51 & 57.53\% ± 0.52 & 39.88\% ± 0.42 \\
86.85\% ± 0.47 & 75.71\% ± 0.49 & \textbf{61.33\% ± 0.41} & 45.56\% ± 0.37 \\
\underline{86.92\% ± 0.54} & \underline{76.07\% ± 0.46} & \underline{61.30\% ± 0.47} & \textbf{45.95\% ± 0.38} \\
\textbf{87.61\% ± 0.47} & \textbf{76.52\% ± 0.44} & 61.23\% ± 0.39 & \underline{45.65\% ± 0.34} \\
\bottomrule
\end{tabular}}
\end{subtable}
\end{table*}
\setlength{\intextsep}{3pt}  
\setlength{\floatsep}{8pt}   
\begin{table*}[h!]
\caption{Hierarchical linear classification on CIFAR-100}
\label{table:5}

\begin{subtable}[t]{0.35\linewidth}
\centering
\caption{\textbf{Pretraining on CIFAR-100}}
\label{tab:cifar100_no_pretrain}
\small
\scalebox{0.62}{
\begin{tabular}{lcccc}
\toprule
\textbf{Model} & \textbf{Level 1} & \textbf{Level 2} & \textbf{Level 3} & \textbf{Level 4} \\
\midrule
\textbf{SimCLR}           & 74.18\% ± 0.42 & 59.31\% ± 0.53 & 41.46\% ± 0.39 & 32.51\% ± 0.34 \\
\textbf{BarlowTwins}      & 76.62\% ± 0.41 & 62.78\% ± 0.43 & 45.31\% ± 0.39 & \underline{36.05\% ± 0.38} \\
\textbf{VICReg}           & \underline{77.23\% ± 0.39} & \underline{63.16\% ± 0.46} & \underline{45.77\% ± 0.41} & 35.55\% ± 0.40 \\
\textbf{SAG-VICReg (Ours)}& \textbf{78.46\% ± 0.44} & \textbf{64.36\% ± 0.42} & \textbf{46.85\% ± 0.43} & \textbf{36.33\% ± 0.36} \\
\bottomrule
\end{tabular}}
\end{subtable}
\hspace{8.5em}  
\begin{subtable}[t]{0.35\linewidth}
\centering
\caption{\textbf{Pretraining on ImageNet-1k}}
\label{tab:cifar100_imagenet_pretrain}
\small
\scalebox{0.62}{
\begin{tabular}{cccc}
\toprule
\textbf{Level 1} & \textbf{Level 2} & \textbf{Level 3} & \textbf{Level 4} \\
\midrule
89.12\% ± 0.51 & 79.75\% ± 0.46 & 68.43\% ± 0.48 & 59.95\% ± 0.51 \\
89.99\% ± 0.41 & 81.78\% ± 0.43 & 71.37\% ± 0.47 & \underline{62.64\% ± 0.48} \\
\underline{90.48\% ± 0.40} & \underline{82.36\% ± 0.42} & \textbf{71.92\% ± 0.44} & \textbf{62.83\% ± 0.35} \\
\textbf{90.79\% ± 0.46} & \textbf{82.80\% ± 0.51} & \underline{71.63\% ± 0.43} & 62.57\% ± 0.46 \\
\bottomrule
\end{tabular}}
\end{subtable}
\end{table*}
\vspace{1em}
\\Tables \ref{table:4} and \ref{table:5} present hierarchical \emph{k}-NN and linear classification results, respectively. These conventional evaluation metrics confirm that SAG-VICReg's enhanced global structure representation does not come at the expense of local discriminative power. Our method achieves the highest accuracy at Level 1 and Level 2 across both classification paradigms, demonstrating its superior ability to capture broad semantic groupings. Remarkably, it also maintains competitive performance at the finest granularity levels (Level 3 and 4), showing only minimal differences compared to the best performers. This dual capability highlights how our approach successfully accounts for both global semantic structure and fine-grained discrimination, achieving an effective balance that benefits hierarchical understanding at all levels.

\section{Conclusion}
In this work, we established the connection between VICReg and spectral embedding, revealing a potential generalization challenge. Our proposed SAG-VICReg addresses this through random-walk pairing, enhancing both stability and global semantic understanding. Experiments demonstrate that our approach effectively captures hierarchical data structure while maintaining strong performance on local metrics. Additionally, we introduced a new \emph{label-free} evaluation metric that better assesses global structure preservation. These contributions advance self-supervised learning by considering both local discrimination and global semantic coherence.
\bibliographystyle{unsrt}
\bibliography{references} 

\newpage
\appendix

\section{Visualizations}
\label{sec:visual}
\begin{figure*}[h!]
    \centering
\includegraphics[width=0.8\textwidth]{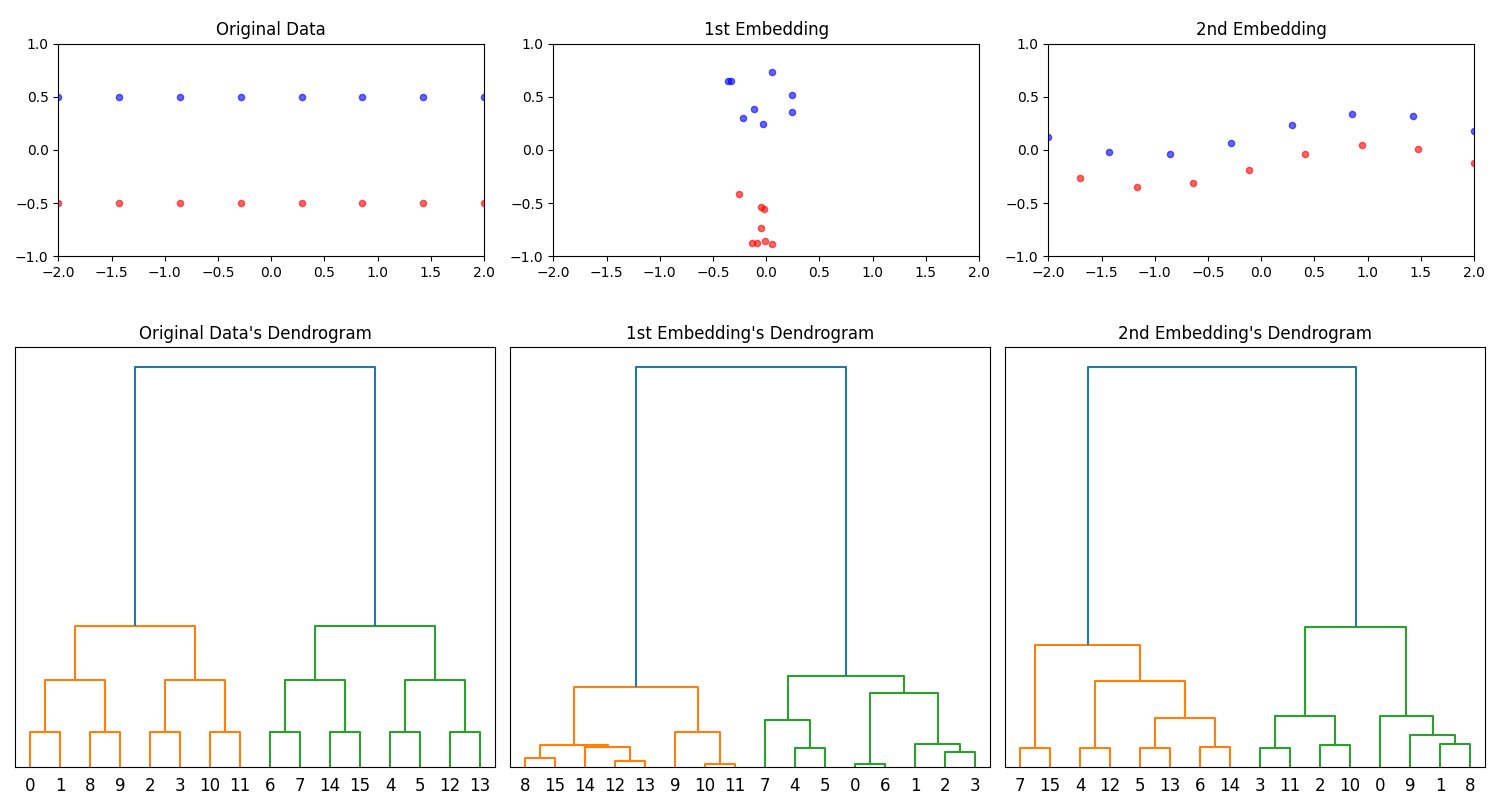} 
    \caption{\textbf{Comparing embedding quality assessment methods}. \textbf{Top:} Original data (left) and two embeddings, where the first (middle) is linearly separable but distorts global structure, while the second (right) preserves structural relationships but isn't linearly separable. \textbf{Bottom:} Dendrograms reveal that the second embedding's hierarchical structure closely matches the original data, maintaining key relationships (e.g., sample 0's proximity to 1, 8, 9) that are lost in the first embedding despite its linear separability. This demonstrates why our metric provides a more comprehensive evaluation than traditional approaches.}
    \label{fig:3}
\end{figure*}

\begin{figure*}[h!]
    \centering
    \includegraphics[width=0.95\textwidth]{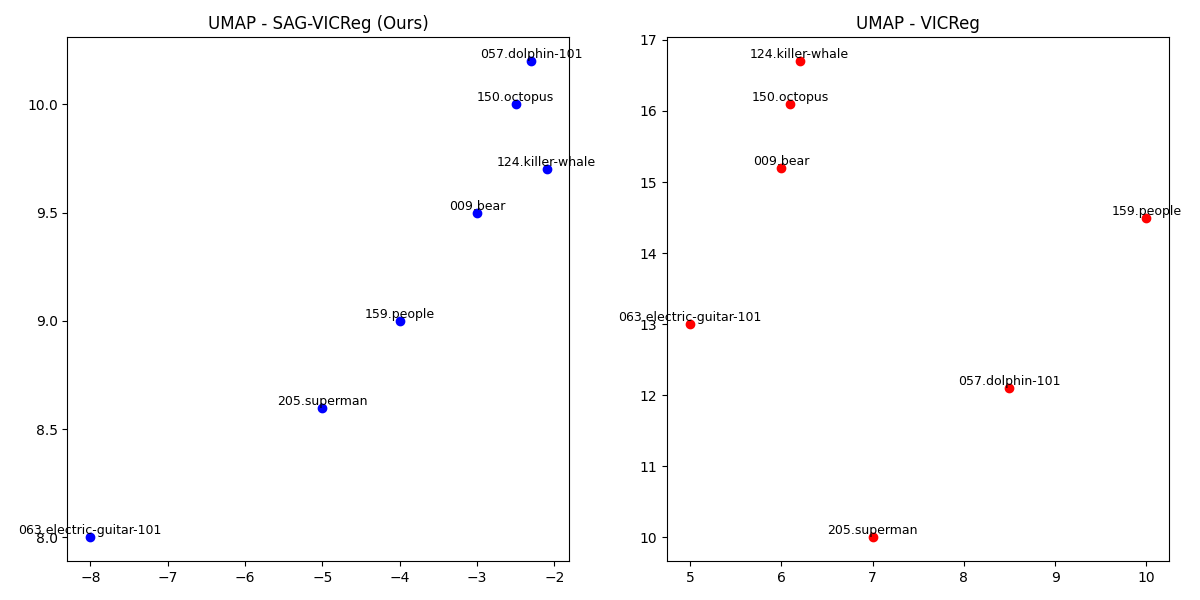}
    \caption{\textbf{UMAP visualizations of the class centroids (mean embeddings) for SAG-VICReg and VICReg.} 
        Both models were trained on the same set of classes, which differ from those used to generate the embedded images. 
        SAG-VICReg more effectively preserves global structure even on unseen data, 
        drawing semantically related centroids closer and pushing distant ones further apart. 
        For example, aquatic classes are more tightly clustered in SAG-VICReg than in VICReg. 
        Additionally, ``superman'' is closer to the ``people'' class centroid, 
        while ``electric guitar'' is placed farther from unrelated classes, 
        showcasing SAG-VICReg's stronger semantic separation compared to VICReg.}
    \label{fig:8}
\end{figure*}

\section{Theoretical justification for our approach}
\label{sec:theory_related}

Our method builds on recent theoretical results that highlight
the centrality of spectral analysis for self-supervised embedding
and generalization. First,
\cite{haochen2021provable} show that in a contrastive
learning framework--modeled by a so-called \emph{augmentation graph},
whose nodes represent augmented datapoints-- richer or denser edges
yield smoother eigenfunctions and more robust embeddings. Concretely,
adding edges for augmentations that capture genuine relationships in data
improves stability for points outside of the original training clusters. 
This insight directly underpins our random-walk sampling strategy in
SAG-VICReg, where we deliberately introduce additional
“positive” edges to foster cross-image relationships (beyond strict
instance-level augmentations). The effect is to densify the augmentation
graph, thereby improving spectral regularization and leading to more stable
embeddings in unseen regions.

Second, \cite{li2023spectral} provide \emph{excess risk bounds} for
spectral clustering and analyze how one can extend the learned embedding
map to newly arrived samples. Their analysis confirms that spectral methods
can reliably embed unseen points from the same manifold, provided that
the embeddings are anchored in sufficiently connected local structures.
In our setting, ensuring that “edge creation” (i.e.\ sampling cross-image
pairs) captures correct semantic neighborhoods helps reduce distortion
when new data, possibly from new clusters, arrives at test time. The
random-walk procedure thus strengthens these local-to-global relationships
and mitigates the generalization gap.

Finally, \cite{cabannes2023ssl} emphasize that creating
nontrivial positive pairs (i.e., carefully chosen augmentations
rather than trivial identity mappings) reduces the variance of the
learned embedding and, more importantly, boosts generalization.
Building on this perspective, our SAG-VICReg strategy makes use of a
\emph{weighted random walk} over the batch’s similarity graph to select
pairs that share broader semantic similarity (e.g., partial overlap or
common neighborhood). Essentially, by sampling from a denser subgraph
under an appropriate random-walk transition, we encourage the model to
recognize higher-level or cross-image relationships, consistent with
the theoretical prediction that such “denser edges” yield better
accuracy on downstream tasks. Hence, our design can be viewed as a
principled approach toward bridging the gap between purely local
augmentation invariance and capturing global manifold structure in the
learned representations.
\section {Hyperparameters}
\label{hyperparameters}
A critical aspect of our SAG-VICReg approach is its stability with respect to hyperparameter choices. Beyond the standard VICReg hyperparameters ($\lambda=25$, $\mu=25$, $\nu=1$), our method introduces two additional parameters: $k$ (the number of neighbors) and \textit{scale} (for the Gaussian kernel).

We determined that $k = 5$ neighbors consistently produces optimal results across various datasets and evaluation metrics. For the scale parameter, we found that using a locally adaptive approach--computing scale as the 20th percentile of the distance matrix after subtracting each sample's nearest-neighbor distance--provides the best balance of performance and stability. This configuration was particularly effective at our standard batch size of 256, but importantly, maintained robust performance even when batch size was varied.

We maintained the original VICReg regularization weights ($\lambda=25$, $\mu=25$, $\nu=1$) and learning rate schedules unchanged, allowing for direct comparison with the baseline approach. This consistency in hyperparameters further demonstrates that SAG-VICReg's performance improvements stem from our architectural innovations rather than parameter tuning.

The stability of SAG-VICReg is evidenced by its consistent performance across our evaluation suite: the newly proposed Hierarchical Rand index (Figure \ref{fig:ri_comparison}), standard classification metrics, and hierarchical linear and $k$-NN classification tests (Tables \ref{table:4}, \ref{table:5}). This robust performance across a reasonable range of $k$ and \textit{scale} values confirms that SAG-VICReg does not require extensive hyperparameter optimization to achieve significant improvements over the baseline.

\section {Architecture}
\label{Arch}
We maintained architectural consistency across our experimental framework to ensure fair comparisons. Our implementation follows the original VICReg architecture as detailed in Figure 1 of \cite{bardes2021vicreg}.
For experiments involving invariance-based methods (Tables \ref{table:1}, \ref{table:2}), we employed dataset-specific backbone configurations. ImageNet and Caltech-256 experiments utilized a ResNet-50 backbone that generated 2048-dimensional representation vectors, subsequently projected to 8192-dimensional embeddings through a series of three fully connected layers. For CIFAR-100, we adapted the architecture to use a ResNet-34 backbone, producing 512-dimensional representations expanded to 1024-dimensional embeddings via three fully connected layers.
Our investigation of masking-based methods (Table \ref{table:3}) employed a consistent ViT-B/16 backbone architecture across all datasets. This configuration generated 768-dimensional representation vectors, which were then projected to 3072-dimensional embeddings through three fully connected layers.
For the hierarchical evaluation experiments (Figure \ref{fig:ri_comparison}, Tables \ref{table:4}, \ref{table:5}), we maintained the same ResNet architectures as used in Tables \ref{table:1} and \ref{table:2}, ensuring that all comparative models utilized identical backbone networks to facilitate direct performance comparison under controlled conditions.
\section{Data}
\label{data}
Our experiments employed a carefully structured approach to data partitioning across multiple datasets.
For the comparisons in Tables \ref{table:1} and \ref{table:2}, we used a consistent split methodology across datasets. We selected the first 100 classes of ImageNet (detailed class names provided in Appendix \ref{classesnames}), dividing them into two equal parts: training the first model on classes 1-50 and the second model on classes 51-100. We applied an identical splitting strategy to CIFAR-100. For Caltech-256, we implemented a 1-128 and 129-256 split. Testing was consistently performed on the first portion of each dataset (classes 1-50 for ImageNet and CIFAR-100, classes 1-128 for Caltech-256).
The experiments in Table \ref{table:3} maintained these same class divisions. To specifically evaluate SAG-VICReg's generalization capabilities, we trained models on the first 50 classes of CIFAR-100 and the first 128 classes of Caltech-256, then tested performance on classes 1-50 of CIFAR-100.
For the Rand Index comparisons (Figure \ref{fig:ri_comparison}) and the hierarchical classification experiments (Tables \ref{table:4}, \ref{table:5}), we utilized the complete CIFAR-100 dataset when pretraining on CIFAR-100, and the full ImageNet-1K dataset when pretraining on ImageNet-1K to provide comprehensive evaluation across the entire data distributions. 
\section{Training details}
\label{setup}
We implemented a comprehensive training protocol tailored to each dataset and model architecture to ensure optimal performance while maintaining experimental validity.
For the invariance-based model comparisons (Tables \ref{table:1}, \ref{table:2}), we trained ImageNet models for $300$ epochs with a batch size of $256$ and a base learning rate of $1.3$. CIFAR-100 and Caltech-256 models underwent more extensive training for $500$ epochs, maintaining the same batch size of $200$. All invariance-based models utilized the LARS optimizer \cite{you2017large, goyal2017accurate} with a weight decay of $10^{-6}$, implementing a cosine learning rate decay schedule \cite{loshchilov2017sgdr} that started from $0$ and incorporated $10$ warmup epochs, consistent with the original VICReg implementation.

For the masking-based methods comparison (Table \ref{table:3}), we trained all models for $250$ epochs using the AdamW optimizer \cite{loshchilov2017decoupled}. We adapted batch sizes according to model requirements: $140$ for I-Jepa, $120$ for DINO, and $200$ for MAE, SAG-VICReg and VICReg. All models employed a cosine learning rate decay schedule \cite{loshchilov2017sgdr} with $10$ warmup epochs. Weight decay parameters were set at $0.05$ for VICReg, SAG-VICReg, and MAE, and slightly lower at $0.04$ for DINO and I-Jepa.
Models used in the Rand Index comparisons (Figures \ref{fig:6}, \ref{fig:7}) were trained for $500$ epochs, for the hierarchical classification experiments (Tables \ref{table:4}, \ref{table:5}), we trained CIFAR-100 models for $500$ epochs while using $120$ epochs for ImageNet.
 
\section{Additional Preliminaries}
\label{sec:preliminaries}
\subsection{Spectral Embedding}
\label{spectralembedding}
Spectral embedding transforms high-dimensional data into a lower-dimensional representation by exploiting the eigenstructure of a graph's Laplacian matrix. This technique preserves the underlying relationships between data points through their graph representation.
Consider a weighted, undirected graph $G = (V, E)$ with $n$ vertices. The graph structure is characterized by two fundamental matrices:
 The weight matrix $W \in \mathbb{R}^{n \times n}$, where:
$W_{ij} = \text{similarity between vertices } i \text{ and } j $.
The degree matrix $D \in \mathbb{R}^{n \times n}$, a diagonal matrix where:
    $ D_{ii} = \sum_{j=1}^n W_{ij} $
The graph Laplacian matrix $L$ is then defined as:
$ L = D - W $.

To compute the $k$-dimensional spectral embedding, we first need to find the eigendecomposition of $L$. Next, from this decomposition, we take the $k$ eigenvectors ${v_1, \ldots, v_k}$ that correspond to the $k$ smallest non-zero eigenvalues. Finally, we construct the embedding matrix $X \in \mathbb{R}^{n \times k}$, where each row $i$ contains the embedded coordinates for vertex $i$ in the lower-dimensional space.
This process transforms each vertex from our original graph into a point in a $k$-dimensional space, where vertices that were strongly connected in the original graph remain close to each other in this new representation.

A major limitation of spectral embedding is that adding new points to the graph requires recalculating \( W \), \( D \), \( L \).
When a new point is introduced, not only do we need to calculate its similarities to all existing vertices, but we also need to update the graph's structure to incorporate these new relationships.
Therefore, finding the spectral embeddings requires a recalculation of the eigen-
decomposition to obtain the new k-dimensional embeddings.
This limitation shows why spectral embeddings are intended primarily for fixed graphs, as in \cite{bengio2004out}, \cite{fowlkes2004spectral} and \cite{coifman2006geometric}.
SpectralNet \cite{shaham2018spectralnet}, aims to solve this generalization issue using Deep Neural network that learns the spectral embedding.

\subsection{Random Walk on a Graph}
\label{randomwalk}
A random walk on a Graph is a stochastic process where each step involves moving from one node to another according to edge weights, capturing the connectivity structure of the graph. Given the weight matrix \( W \) and degree matrix \( D \), the random walk matrix \( P \) is defined as:
$P = D^{-1} W$,
where \( P_{ij} \) represents the probability of transitioning from node \( i \) to node \( j \) in a single step:
$
P_{ij} = \frac{W_{ij}}{D_{ii}}
$
This matrix \( P \) is row-stochastic, meaning each row sums to 1, representing the transition probabilities of a Markov chain. To sample an index \(j\) from a Random walk matrix \(P\) in the \(i\)-th row, we choose one index according to the probability distribution defined by \(Pij\).

\subsection{Cophenetic Correlation}
\label{cope}
Cophenetic correlation \cite{cope1962} is a metric used to evaluate how well a hierarchical clustering structure preserves the pairwise distances between observations in a dataset. This measure is particularly useful in embedding evaluation, as it assesses whether embeddings reflect the hierarchical relationships inherent in the data.
Given a hierarchical clustering, the cophenetic distance between any two points \( i \) and \( j \) is defined as the height of the dendrogram at which these two points are first joined into a single cluster. Let \( T(i, j) \) denote this cophenetic distance for points \( i \) and \( j \) based on the hierarchical clustering.
The cophenetic correlation coefficient \( c \) quantifies the correlation between the original pairwise distances \( D(i, j) \) in the data and the cophenetic distances \( T(i, j) \). Mathematically, it is calculated as:
\[
c = \frac{\sum_{i < j} \left( D(i, j) - \bar{D} \right) \left( T(i, j) - \overline{T} \right)}{\sqrt{\sum_{i < j} \left( D(i, j) - \bar{D} \right)^2 \sum_{i < j} \left( T(i, j) - \overline{T} \right)^2}}
\]
where:
\( \bar{D} \) is the mean of all pairwise distances \( D(i, j) \),
and \( \overline{T} \) is the mean of all cophenetic distances.
The distance metric of The distance metric of \(D\) can be Euclidean or cosine distance, as well as other metrics.
The cophenetic distance \(T\) can be calculated by tracing the dendrogram tree and identifying the height at which clusters merge.

\subsection{Rand index}
\label{rand_index}
The Rand Index \cite{rand1971objective} is a measure for quantifying the similarity between two clusterings of the same dataset.
Let there be a dataset \(\{x_1, x_2, \dots, x_n\}\) of \(n\) elements, and suppose we have two partitions (clusterings) of this dataset, denoted by \(X\) and \(Y\). Each partition assigns every element \(x_i\) to a cluster label. The Rand Index measures how consistently pairs of elements are assigned to the same cluster or to different clusters in both partitions.

Formally, consider all possible \(\binom{n}{2}\) pairs of distinct elements. A pair \((x_i, x_j)\) is counted as an \emph{agreement} if they are either:
\begin{itemize}
    \item Placed in the same cluster under both partitions \(X\) and \(Y\), or
    \item Placed in different clusters under both partitions \(X\) and \(Y\).
\end{itemize}
The Rand Index is given by the fraction of such pairwise agreements out of all possible pairs:
\[
\text{RandIndex}(X, Y) = \frac{\text{number of pairwise agreements}}{\binom{n}{2}} .
\]
A Rand Index close to \(1\) indicates a high level of agreement between the two clusterings, while a value closer to \(0\) indicates lower agreement.

\section{Rand index setup}
\label{rand_index_setup}
In this configuration, all models are trained on the full CIFAR-100 dataset for 250 epochs, and unlike the LCA and Cophenetic similarity approaches, we do not train two separate models per method. We begin by constructing a hierarchical tree of the CIFAR-100 label structure, following the hierarchy described by \cite{kim_class_hierarchies} (Table S1). For each level in this hierarchy, we compute the Rand index by comparing two partitions: (1) the clustering assignments obtained via spectral clustering on the embeddings, and (2) the corresponding labels at that level. The number of clusters is set to match the number of classes at each level.

Beyond this hierarchical approach, we also calculate the Rand index at the finest level of granularity (i.e., all 100 CIFAR-100 classes). In this case, the number of clusters produced by spectral clustering is allowed to vary, rather than being fixed to the total number of classes.

The corresponding results for these approaches are presented in Figures \ref{fig:6} and \ref{fig:7}, respectively.

\section{Implementation details}
\label{implementation}
The affinity matrix \( W \) is computed to represent the relationships between points in \( X \) and \( Y \). The steps are as follows:

\subsection{Nearest Neighbors and Distance Adjustment}
The distances to the \( k \)-nearest neighbors (\( \text{D} \)) are computed using cosine distance, where \( k = 5 \). To emphasize relative relationships, the distance to the nearest neighbor is subtracted from all distances, inspired by \cite{McInnes2018}:
\[
\text{D}_{ij} = \text{D}_{ij} - \text{D}_{i0}, \quad \forall j
\]

\subsection{Scale Calculation}
The scale parameter determines the Gaussian kernel width and is derived based on the adjusted distances (after decreasing the distance to the nearest neighbor from the distance to every sample point, which results in nearest distance \(=0\)). For local scaling, the 20th percentile of the adjusted distances is used:
\[
\text{scale}_i = \max(\text{Percentile}(\text{D}_i, 20), 1e^{-7})
\]
The \( 1e^{-7} \) ensures numerical stability by clamping the scale to a minimum value.

\subsection{Gaussian Kernel}
Using the computed scale, the Gaussian kernel is applied to the pairwise distances \( D \):
\[
W_{ij} = \exp\left(-\frac{D_{ij}^2}{\text{scale}_i^2}\right)
\]
This creates the affinity matrix, emphasizing points within the local neighborhood.

\subsection{Masking Non-Neighbors}
A binary mask is applied to sparsify \( W \), keeping only the entries corresponding to the \( k = 5 \) nearest neighbors:
\[
W_{ij} = 
\begin{cases} 
W_{ij}, & j \in \text{Ids}_i \\
0, & \text{otherwise}
\end{cases}
\]

\section{Hardware and Computational overhead}
\label{hardware}
We use NVIDIA L40S for our evaluations, and NVIDIA GeForce GTX 1080 Ti as well the L40S for our trainings.
We benchmarked \textsc{SAG-VICReg} versus \textsc{VICReg} on an NVIDIA~L40 GPU,
using the full ImageNet~1k dataset (\(1{,}281{,}167\) samples per epoch). 
\begin{itemize}
\item \textbf{Runtime per epoch:} 
    \textsc{SAG-VICReg}: 3600s, 
    \textsc{VICReg}: 2800s.
\item \textbf{GPU memory:}
    \textsc{SAG-VICReg}: 24,389\,MB,
    \textsc{VICReg}: 24,194\,MB.
\end{itemize}

Overall, \textsc{SAG-VICReg} incurs a \(\sim 28.5\%\) increase in 
computational time, with nearly the same memory overhead as \textsc{VICReg}.
These results confirm that \textsc{SAG-VICReg} remains scalable and 
feasible for large datasets.

\section{ImageNet classes}
\label{classesnames}
The following are the first 100 classes of ImageNet used in our experiments:

\noindent
n01440764, n01443537, n01484850, n01491361, \\
n01494475, n01496331, n01498041, n01514668, \\
n01514859, n01518878, n01530575, n01531178, \\
n01532829, n01534433, n01537544, n01558993, \\
n01560419, n01580077, n01582220, n01592084, \\
n01601694, n01608432, n01614925, n01616318, \\
n01622779, n01629819, n01630670, n01631663, \\
n01632458, n01632777, n01641577, n01644373, \\
n01644900, n01664065, n01665541, n01667114, \\
n01667778, n01669191, n01675722, n01677366, \\
n01682714, n01685808, n01687978, n01688243, \\
n01689811, n01692333, n01693334, n01694178, \\
n01695060, n01697457, n01698640, n01704323, \\
n01728572, n01728920, n01729322, n01729977, \\
n01734418, n01735189, n01737021, n01739381, \\
n01740131, n01742172, n01744401, n01748264, \\
n01749939, n01751748, n01753488, n01755581, \\
n01756291, n01768244, n01770081, n01770393, \\
n01773157, n01773549, n01773797, n01774384, \\
n01774750, n01775062, n01776313, n01784675, \\
n01795545, n01796340, n01797886, n01798484, \\
n01806143, n01806567, n01807496, n01817953, \\
n01818515, n01819313, n01820546, n01824575, \\
n01828970, n01829413, n01833805, n01843065, \\
n01843383, n01847000, n01855032, n01855672.

\end{document}